\documentclass{amia}
\usepackage{lipsum} %Remove if not needed
 
\usepackage[load-configurations=version-1]{siunitx} % newer version
\usepackage{multirow}

\usepackage{booktabs}
\usepackage{longtable}
\usepackage{xspace}
\usepackage{xcolor}
\usepackage{enumitem}
\usepackage{subfigure}
\usepackage{hyperref}

\newcommand\lucy[1]{{\color{red}\{\textit{#1}\}$_{lucy}$}}

\newcolumntype{L}[1]{>{\raggedright\let\newline\\\arraybackslash\hspace{0pt}}p{#1}}
\newcolumntype{M}[1]{>{\raggedright\let\newline\\\arraybackslash\hspace{0pt}}m{#1}}
\newcolumntype{C}[1]{>{\centering\arraybackslash\hspace{0pt}}p{#1}}

\definecolor{darkgreen}{rgb}{0.0, 0.4, 0.13}
\definecolor{purple(munsell)}{rgb}{0.62, 0.0, 0.77}

\begin{document}

\title{Explainable AI for Clinical Outcome Prediction: A Survey of Clinician Perceptions and Preferences}

% \author{\Name{Jun Hou}
%        \Email{hou.jun@northeastern.edu}\\ 
%        \addr Department of Mechanical and Industrial Engineering\\
%        Northeastern University\\
%        Seattle, Washington State, USA 
%        \AND
%        \Name{Lucy Lu Wang}
%        \Email{lucylw@uw.edu}\\ 
%        \addr Information School\\
%        University of Washington\\
%        Seattle, Washington State, USA} 

\author{Jun Hou, MS$^1$, Lucy Lu Wang, PhD$^2$ }

\institutes{
    $^1$Virginia Tech, Blacksburg, VA; $^2$University of Washington, Seattle, WA
}

\maketitle

\begin{abstract}
  % Outcome prediction in clinical settings can help identify patients at higher risk of negative outcomes, such as treatment failure or in-hospital mortality.
  % With deep learning-based models in particular, 
  Explainable AI (XAI) techniques are necessary to help clinicians make sense of AI predictions and integrate predictions into their decision-making workflow.
  % Our aim is to 
  % understand how clinicians assess the outputs of XAI techniques that can be applied to clinical text.
  % Specifically, we focus on 
  In this work, we conduct a survey study to understand 
  % which XAI techniques can help clinicians interpret predictions over text-based EHR data, 
  clinician preference among different XAI techniques when they are used to interpret model predictions over text-based EHR data.
  % , and how these preferences might guide development of better XAI techniques.
  We implement four XAI techniques (LIME, Attention-based span highlights, exemplar patient retrieval, and free-text rationales generated by LLMs) on an outcome prediction model that uses ICU admission notes to predict a patient's likelihood of experiencing in-hospital mortality.
  % ; the techniques we investigate include LIME, Attention-based span highlights, exemplar patient retrieval, and free-text rationales generated by a large language model. 
  Using these XAI implementations, we design and conduct a survey study of 32 practicing clinicians, collecting their feedback and preferences on the four techniques. We synthesize our findings into a set of recommendations describing when each of the XAI techniques may be more appropriate, their potential limitations, as well as recommendations for improvement. 
\end{abstract}

\section*{\textsc{Introduction}}
\vspace{-1mm}
% \jun{There is no option to load supplementary materials.}
Clinical decision support systems (CDSS) powered by machine learning and AI have the potential to assist in medical decisions and improve patient outcomes. 
% Applications of CDSS have been explored in a variety of clinical settings, such as the Intensive Care Unit (ICU),\cite{Berge_2023} Oncology,\cite{Shen2017ConstructingOC} and Cardiology. \cite{kim2022prediction} 
However, to meaningfully support clinicians, AI-powered CDSS must be trustworthy and interpretable, allowing clinicians to assess the utility and applicability of model predictions. Explainable AI (XAI) techniques have been proposed to improve model interpretability, especially for neural network and other blackbox models.\cite{danilevsky2020survey} While XAI techniques have been applied to CDSS,\cite{feng2020explainable} a comprehensive understanding of clinician preferences and perceptions regarding XAI applications in these systems remains largely unexplored. 

% Researchers have rarely followed up with the end-users of these models (clinicians) regarding the usefulness of XAI techniques applied on top of predictions for assisting with clinical decisions.
% in real-world settings. 
% Reviews and survey papers in this space 
Prior work on clinical XAI tends to focus on explanatory accuracy, in terms of which models are applicable,\cite{Sheikhalishahi_2019} how to integrate XAI methods for different healthcare tasks,\cite{Chaddad_2023} or which datasets are available to train on.\cite{di2023explainable} 
%Efforts exploring clinician feedback in this space have also focused more on Computer Vision applications \cite{Chen2021ExplainableMI} rather than Natural Language Processing (NLP) applications, which are receiving more attention due to recent advances in language modeling.
% Recent studies have also applied Large Language Models (LLMs) with some success to healthcare tasks such as digital medicine\cite{tsai2022natural} and radiology report generation.\cite{Moezzi_2022}
While these works consider XAI for improving model interpretability, they do not incorporate user studies with clinical practitioners to understand whether XAI methods meet their needs. Solely focusing on technical performance metrics can lead to a gap between how AI developers and healthcare professionals view and assess AI tools.
% \jun{Addressing why we focus on opinion} 

% Though researchers and practitioners seem to agree on the need for XAI in healthcare, there is limited validation with end users regarding the design and appropriateness of these methods in clinical settings.
%In this work, we aim to fill this gap. 
% who spend the most time with patients and will benefit most from XAI-enhanced CDSS models that improve communication between patients and doctors.
% Here, we aim to understand how clinicians perceive different XAI techniques, whether they are judged to be useful, and how each technique can be improved to facilitate clinician understanding of model output and support their decision-making workflow.
We fill this validation gap by surveying clinical workers about the utility of popular XAI methods applied to text-based CDSS. We conduct a study with 32 clinicians (predominantly nurses along with physicians, technicians, and administrators), asking them to interact with and compare several XAI methods, and eliciting feedback about the design and utility of these methods. 
% Based on a review of XAI usage in clinical decision support, 
We  methods in our study, including LIME (Local Interpretable Model-Agnostic Explanation),\cite{Ribeiro2016WhySI} Attention-based span highlights,\cite{Vaswani2017AttentionIA} exemplar patient retrieval, and free-text rationales produced by a large language model (LLM), which represent all explanation forms categorized in the review by Chaddad et al.\cite{Chaddad_2023}
We apply these methods to explain outputs for a model trained to predict in-hospital mortality in an intensive care setting, a task studied extensively in prior work.\cite{, van-aken-etal-2021-clinical, jin2018improving} We select this task because it mirrors decision-making processes common in disease diagnosis, supporting the potential to generalize our findings to other similar scenarios.
% where binary classification is applicable. 
% \jun{Addressing the single MOR task}
% randl2023early, Mondrejevski2022FLICUAF, 
%, bardak2020improving, huang2019patient
% We design a questionnaire based on our predictive model and XAI techniques to investigate clinician perceptions and preferences for these techniques. 
% We survey 
Based on participants' questionnaire results, we answer the following research questions: (1) how these XAI methods can be improved, (2) what tasks they potentially support, and (3) how they compare. 

After conducting thematic analysis on the results, we synthesize a set of recommendations for 
% how researchers and practitioners should 
% approaching 
designing and implementing XAI methods in an ICU setting. 
Our findings underscore the importance of creating both efficient, generalized tools and specialist-sensitive options tailored to varying levels of clinical expertise. We also observe a strong preference among clinicians for free-text rationales, highlighting their potential to enhance communication between healthcare providers and patients. However, our participants also emphasize the importance of evidence-based XAI approaches, such as similar patient retrieval, in building trust between clinician users and AI systems.

%\todoit{summarize insights below in text and remove section}

    %\item Clinician feedback is essential for designing XAI methods that meet the real-world needs of clinical settings. We design and conduct a survey around four representative XAI methods    to capture the perceptions and preferences of clinicians.
    %\item Our findings underscore that not every clinician has the same need for XAI tools. However, clinician experience and workflow can be influential on their acceptance and preferences, which suggests that a generalized XAI method that optimizes efficiency and clarity should be coupled with a specialist-sensitive tool to support decision-making.
    %\item A preference for free-text rationales is clear among clinician feedback. Its potential as a communication tool can bridge the gap between different healthcare providers and between providers and patients. Also, we find that evidence-based XAI methods such as similar patient retrieval may enhance trust.

\section*{\textsc{Background \& Related Work}} \label{sec:background}
\vspace{-1mm}

\paragraph{Explainable AI (XAI) in healthcare} 
% Chaddad et al.~2023\cite{Chaddad_2023} provide a framework for categorizing XAI methods by four criteria: explanation form, interpretation type, model specificity, and explanation scope. 
% Multiple XAI methods can be categorized as 
Feature-map methods such as LIME\cite{Ribeiro2016WhySI}
% used in \cite{garg2023lonxplain,garg2023annotated} and \cite{Saxena2022ExplainableCA}, 
and SHAP (SHapley Additive exPlanations) \cite{lundberg2017unified}
have been explored repeatedly in clinical settings,\cite{garg2023annotated,Saxena2022ExplainableCA}
% used in \cite{garg2023annotated} 
% Thorsen et al.~2022,\cite{thorsen2022discrete} 
% and attention-based methods.\cite{vaswani2017attention}
% used in \cite{franz2020deep}. 
% On the other hand, 
as have textual explanation forms.\cite{Agerri2023HiTZAntidoteAE, Wang2023CanLL}
% have also been used to explain medical decisions to end-users
% and to explain the rationales behind dementia diagnosis through cognitive tests.\cite{Wang2023CanLL} 
Shen et al.\cite{shen2018constructing}~applied an example-based XAI method, which retrieves similar clinical cases through Case-Based Reasoning. 
In clinical settings, the focus is on post-hoc local explanations that balance accuracy and clarity and provide detailed insights for individual patient cases.\cite{Markus_2021}
Therefore, we focus on post-hoc instance-level explanations in this work,
comparing four XAI methods that span the explanation forms discussed in Chaddad et al.\cite{Chaddad_2023} 

\paragraph{In-hospital Mortality Prediction}
In-hospital mortality prediction aims to estimate the risk of a patient dying during their hospital stay, and is crucial for prioritizing treatment strategies and resource allocation. Prior studies have investigated this task in the ICU setting using clinical text and time series Electronic Health Record (EHR) data.\cite{jin2018improving,Marafino_2018}
% Jin et al.\cite{jin2018improving}~expanded the scope of this task by proposing a multimodal neural network that integrates time-series signals with clinical text. 
%Models that do not use textual data have also been used to improve in-hospital mortality prediction, such as community-based federated learning algorithms \cite{huang2019patient} and convolution-based models \cite{bardak2020improving}. 
Performance on the task was significantly enhanced by leveraging LLMs in Van Aken et al.\cite{van-aken-etal-2021-clinical} Naik et al.\cite{naik-etal-2022-literature}~then integrated patient-specific retrieved literature as input into predictive models to enhance performance.  
% Recent work explored further the use of federated learning for mortality prediction in the ICU setting and its early prediction to assist in treatment planning.\cite{Mondrejevski2022FLICUAF, randl2023early} 
These works highlight the continuing focus on this critical task using diverse methodologies, driving our choice of this task.

\paragraph{Understanding Physician Perspectives and Preferences}
% \jun{should we shorten this paragraph? We can integrate more details of survey into introduction?}
A literature review conducted by Antoniadi et al.\cite{antoniadi2021current} revealed the importance of XAI in building trustworthy AI/machine learning-based CDSS, as well as the lack of user studies in their development. 
%\citet{Bienefeld_2023} explored the gap between clinician needs and developer objectives in Neuro ICUs, and \citet{Wysocki_2023} investigated the communication gap between healthcare professionals and AI models designed for COVID-19 diagnosis.
% \citet{Bienefeld_2023} and \citet{Wysocki_2023} identified gaps between clinician needs and developer objectives in Neuro ICUs and communication in AI models for COVID-19 diagnosis, respectively.
% Some work has incorporated limited user studies in the development process of explainable CDSS, though mostly for medical image or sensor data.
% For example, \citet{born2021accelerating} used Class activation mapping (CAM) to explain lung ultrasound images, which was validated by two clinical experts. \citet{neves2021interpretable} investigated three local model-agnostic XAI methods for the classification of arthythmia through ECG image data; in their investigation, they conducted a small user study with three ECG readers to evaluate the effectiveness of these methods for improving clinicians' classification accuracy. 
% While these works are aligned with ours in employing user studies to evaluate the effectiveness of XAI for improving CDSS, they are limited to specific XAI methods, as well as applications only on image, signal, or descriptive data for specialized predictive tasks. 
A systematic review conducted by Jung et al.\cite{Jung2023EssentialPA} also found that prior work on XAI in healthcare lacked a consensus evaluation framework for assessing the success of the XAI method. We hope to approach these recommendations from a user-centered perspective, based on what clinicians identify as useful aspects of XAI applied in Natural Language Processing (NLP)-powered CDSS. 

\section*{\textsc{Materials \& Methods}} \label{sec:Methods}
\vspace{-1mm}
% We design and conduct a survey study to investigate clinician perceptions of XAI methods in a clinical NLP setting. 

\paragraph{Data \& Task} \label{sec:data}
The data for our study is derived from MIMIC-III,\cite{Johnson_2016} a collection of de-identified health records from 46,520 patients who stayed in the critical care units of Beth Israel Deaconess Medical Center between 2001-2012.
We adopt the early-detection mortality prediction task from Van Aken et al.,\cite{van-aken-etal-2021-clinical} which uses patient admission notes to predict whether a patient will experience in-hospital mortality. Each patient's admission note is semi-structured free text, consisting of the sections Chief complaint, Present illness, Medical history, Admission Medications, Allergies, Physical exam, Family history, and Social history. We use train/test splits from their work,\cite{van-aken-etal-2021-clinical} which consist of 30,420 patients in the
survived class and 3,534 patients in the mortality class in the train split; and 8,797 patients
in the survived class and 1,025 patients in the mortality class in the test split.

\paragraph{Predictive Model} \label{sec:Predict}
We train predictive models on mortality prediction and select examples for explanation generation and inclusion in our survey.
We use UmlsBERT \cite{michalopoulos-etal-2021-umlsbert} as our base model because it was found to be the most effective for in-hospital mortality prediction in prior work.\cite{naik-etal-2022-literature}
UmlsBERT is a semantically-enriched model 
% that initializes from a pretrained 
% BioClinicalBERT model \cite{alsentzer-etal-2019-publicly} 
% clinical BERT model and further 
pretrained on MIMIC-III and the UMLS Metathesaurus; a single linear layer is then added 
for adaptation to downstream classification tasks. In our case, we finetune the model on mortality outcomes from MIMIC-III
following the non-literature-augmented model variant introduced by Naik et al,\cite{naik-etal-2022-literature} achieving 87.86 micro-F1 and 66.43 macro-F1 on a held-out test set.
% \footnote{We could not replicate the results obtained by \cite{Naik2021LiteratureAugmentedCO} exactly, but all differences were within $\pm$0.5 points F1.}

\paragraph{Implementation of XAI Methods} \label{sec:XAI_implementation}

We apply post-hoc XAI methods to our model predictions to create explanations. We experiment with 
% XAI methods representing all explanation forms categorized by \cite{Chaddad_2023}, including 
feature map, textual, and example-based explanation forms,\cite{Chaddad_2023} specifically LIME,\cite{Ribeiro2016WhySI} Attention-based explanations,\cite{Vaswani2017AttentionIA} exemplar explanations through similar patient retrieval, and free-text rationales from an LLM:

\begin{figure}[t!]
    \centering
    \includegraphics[width=\textwidth]{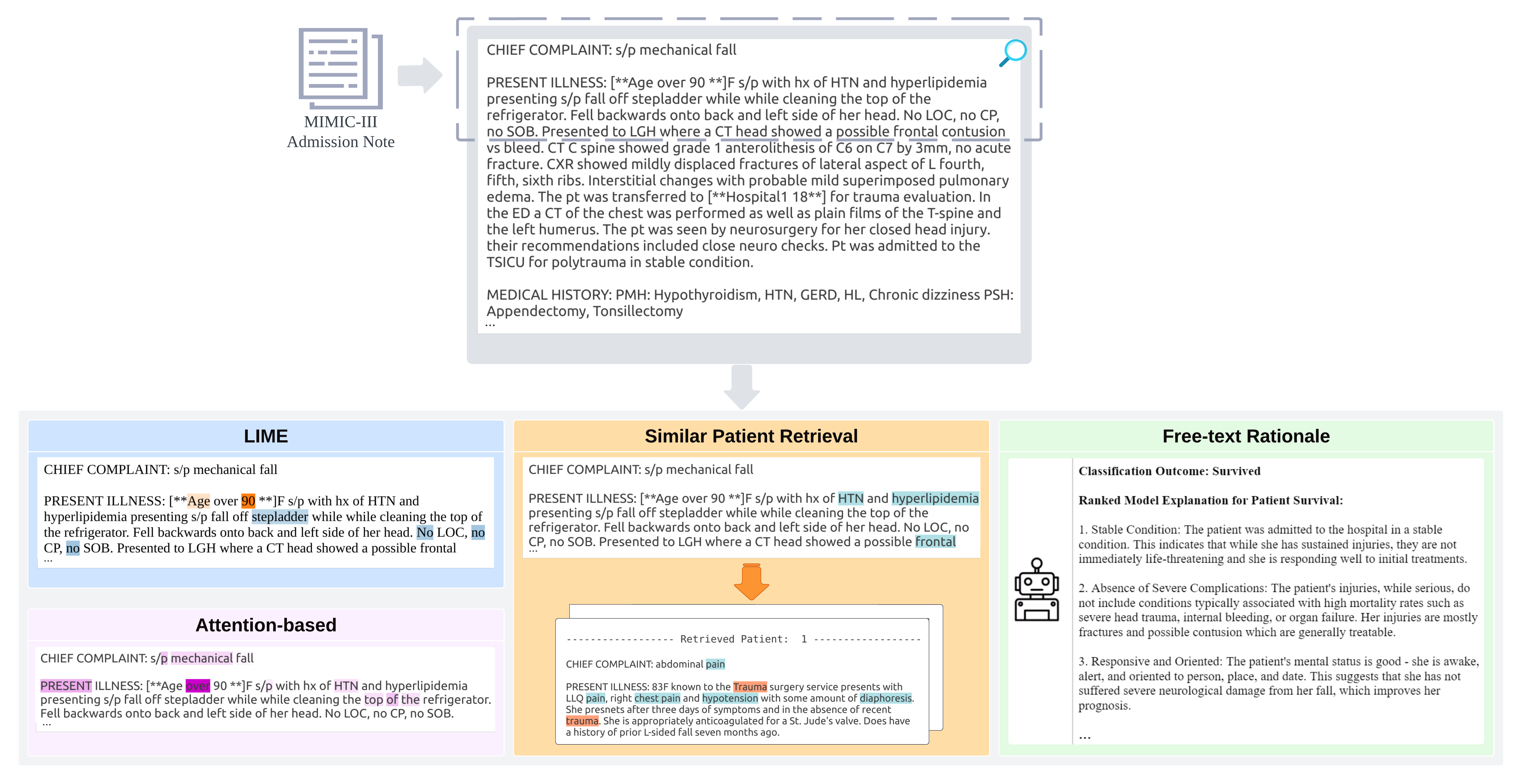}
    \caption{Example outputs of the four XAI methods applied to an MIMIC-III admission not, with color code and intensity indicating feature importance as detailed in the respective sections.
    }
    \label{fig:XAI_samples} 
\end{figure}

\begin{itemize}[topsep=1pt, itemsep=1pt, leftmargin=10pt]
    \item \textbf{LIME} \cite{Ribeiro2016WhySI} is a model-agnostic XAI method that provides feature-based explanations. LIME perturbs the input data by altering or removing features and observes corresponding changes in the model's prediction. Following LIME, we train an explainer module that treats each word in the input document as a feature, and identifies relevant features in a given patient’s admission note that contribute to the prediction result for each of the two outcome classes: mortality (positive) and survived (negative). We present these explanations by highlighting words identified as relevant features for each class, with the intensity of the highlight reflecting the magnitude of the feature's importance. We introduce a percentile variable to limit the number of highlighted words to only the top percent of features based on their importance scores.
    \item \textbf{Attention-based} explanations use weights from attention-based models \cite{Vaswani2017AttentionIA} to explain model decisions, making it a model-specific method. We apply the approach outlined by Falaki et al.\cite{falaki2023attention}~to extract attention weights from our UmlsBERT model's [CLS] tokens and recombine subword tokens into words for visualization. We highlight the top 30\% of words by attention weights, with the intensity of the highlight defined by the weight value.
    \item \textbf{Similar Patient Retrieval} is an exemplar-based, model-agnostic XAI method aimed at producing explanations by retrieving the closest examples from a training dataset. We fine-tune UmlsBERT on mortality prediction and semantic textual similarity through contrastive learning following the SentenceBERT framework.\cite{Reimers2019SentenceBERTSE} We embed all patients with this finetuned model, then apply $k$-nearest neighbor retrieval using cosine similarity to identify similar patients. At inference time, we retrieve the top-3 most similar patients from the training split with the same outcome as what is predicted to show as exemplars. To facilitate visual comparison of similarities and differences between retrieved patients and the test patient, we apply Named Entity Recognition (NER) using scispaCy\cite{neumann-etal-2019-scispacy} and highlight matching entities between test and retrieved notes in orange and non-matching entities in blue. 
    \item \textbf{Free-text Rationales} are a model-agnostic XAI method that attempt to generate human-comprehensible explanations in natural language. Recently, this has often been achieved by prompting LLMs such as GPT-4,\cite{openai2024gpt4} as we do in this work. Our prompts can be found in our Github. We sample six admission notes and their ground-truth labels from the train split based on the method proposed in Liu et al.\cite{liu-etal-2022-makes} for use as in-context examples. We then present the test note and ask for the top 3 reasons for and against the predicted outcome label.
\end{itemize}

Several of these methods are \emph{model-agnostic}, i.e., explanations are generated by a separate model than the one making the prediction. These methods leverage surrogate models and perturbation techniques to approximate prediction model behavior; because the mechanism of the explanatory model differs from the prediction model for these methods, researchers have questioned the fidelity of their explanations.\cite{danilevsky2020survey} For this reason, we also include the \emph{model-specific} method (Attention-based explanations) in our investigation.
 
Figure~\ref{fig:XAI_samples} shows examples of the four XAI methods applied to an example patient admission note. In each case, we give the admission note, model prediction, and in some cases the model itself, as input into the XAI method. We then conduct additional postprocessing of the results to facilitate visualization and comparison. Details and code implementations for all methods can be found on Github: \href{https://github.com/JuneHou/XAI_MOR_Survey.git}{https://github.com/JuneHou/XAI\_MOR\_Survey.git} 

\paragraph{Survey Design} 
\label{sec:Survey_Design}
% The objective of our survey is to explore clinical practitioners' perceptions and preferences towards XAI techniques when applied to explain AI-powered decision support. 
To explore clinician perceptions of XAI methods, we design a survey eliciting feedback for each of the implemented XAI methods and comparing across them.
% whether and how much each XAI method improves clinician understanding of the predictive model, 
We structure our survey into key sections as shown in Figure~\ref{fig:survey_flowchart}. %After obtaining consent, we first assess the participant's baseline perceptions of AI and its use in clinical medicine. Then, we expose participants to each of the four XAI methods, order randomized, and collect feedback, followed by questions designed to assess preference among the different techniques. Lastly, we collect socio-demographic information. 
Following survey completion, we follow up with a subset of participants
% is invited to join a follow-up online interview 
to better understand their preferences and the rationales behind their answers. 
% In the survey, participants were shown the admission notes from one patient with each outcome drawn from this sample. The order of XAI methods shown in the survey was also randomized.
We describe the design of each survey section below:

\begin{figure}[t!]
    \centering
    \includegraphics[width=\textwidth] {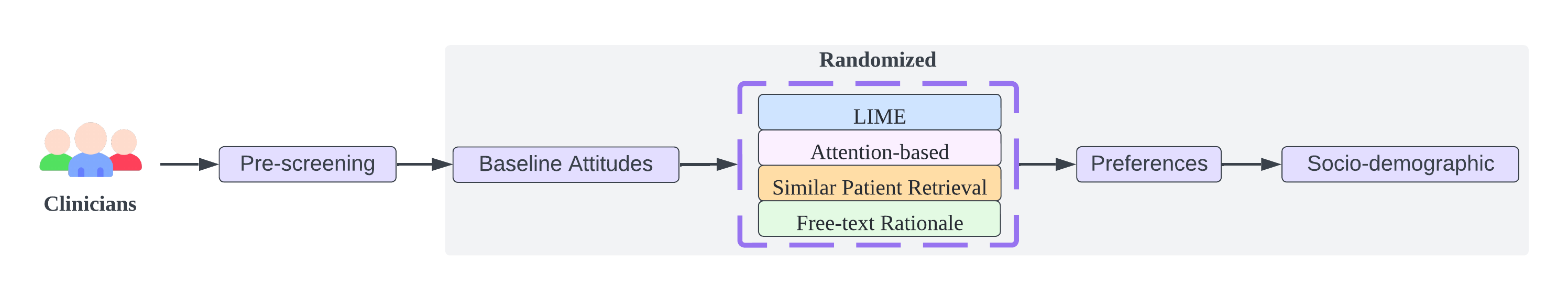}
    \caption{ 
    Survey design and workflow}
    \label{fig:survey_flowchart} 
\end{figure}

\begin{itemize}[topsep=1pt, itemsep=1pt, leftmargin=10pt]
    \item \textbf{Baseline Attitudes}: Personal experiences with AI systems have been found to impact user perceptions of AI.\cite{brauner2023does,kaya2024roles}
    Therefore, we ask participants to choose a set of five values from Jakesch et al.\cite{jakesch2022different} that they consider most important for clinical AI systems and to indicate their attitudes towards AI on a 5-point Likert scale. 
    
    \item \textbf{XAI Perceptions \& Preferences}: We show participants four samples, one of each of the XAI methods with order randomized.
    % , generated as described before. 
    Each XAI technique used in our study is accompanied by a detailed explanation of how it works at the beginning of the respective section.
    % \jun{Address reviewer 2 question regarding lack of insight into the model by healthcare providers}
    To offset patient- and outcome-specific biases, we sample two patients for each of the mortality and survived outcomes to include. 
    % \jun{more details about the randomize of samples to address reviewer 3 question?} 
    We review XAI outputs to ensure that examples shown to users in the survey do not contain obviously incorrect or irrelevant information. 
    
    After each example, the participant answers Likert-scale questions on the understandability, reasonableness, and usefulness of that XAI method; these facets are inspired by the evaluation of explainability and interpretability described by Saeed et al.\cite{saeed2023explainable}~A free-text field is provided to allow the participant to express pros and cons in their own words. For each XAI method, we design additional questions specific to the characteristics of that method, e.g., whether the percent of words highlighted or number of similar patients retrieved are too many or too few. 
    % After answering all questions for one XAI method, the participant moves on to the next method.
    
    Following the evaluation of all four XAI samples, the participant is asked to rank the understandability and reasonableness of all four methods, and indicate their preference among them. We further ask the participant to assess the time efficiency of each method, and whether each method achieved the goals of enhancing confidence, broadening perspective, and increasing trust, three criteria defined in prior work as goals of XAI.\cite{antoniadi2021current}
    \item \textbf{Socio-demographic Information} We collect self-reported gender, age, and race/ethnicity from participants, along with their highest level of education and number of years of clinical experience. Participants also reported their job title, which we categorize into different job positions for reporting.
\end{itemize}

\paragraph{Study Recruitment} \label{sec:recruitment}
% The participants are from two cohorts. 
Participants are eligible if they are
% We screening people working in the domain with the keywords including, healthcare, nurse, doctors. After receiving the proposal, the pre-screening survey is used to find the cohort that is either 
clinical practitioners with more than two years of clinical experience OR medical school students with at least two years of training. We required that all participants be over 18 years of age, have at least a bachelors degree, and are located in the US. 
% For survey participants, we invited physician or nurse who works in ICU setting or surgical and critical care to participant in the follow-up interview. 
% The full description of these cohorts socio-demographic information is shown in Section~\ref{sec:cohort}. 
Participants were recruited using the Upwork platform,
% \footnote{https://www.upwork.com/} 
and compensated at rates of \$15-\$25 per hour for completing pre-screening, the main survey, and any followups. 
% Our recruitment materials can be found in Appendix~\ref{appendix:recruitment}.
%\lucy{put recruitment information here. used upwork and paid \$XX-XX. add recruitment materials to the appendix}
Our study was found exempt by the IRB at the University of Washington (STUDY00019118).

\section*{\textsc{Results}} \label{sec:results}
\vspace{-1mm}
% \subsection*{Cohort Description} \label{sec:cohort}

We recruited 32 clinical practitioners to participate in our study. 
%Participant demographics and work information are provided in Table~\ref{table:dem_info}. 
A majority of participants identified as white (75\%), followed by hispanic/latino/a/x (12.5\%), South Asian (6.3\%), African-American/Black (3.1\%), and Southeast Asian (3.1\%). 78.1\% of participants identified as female. Participant distribution across age groups is more balanced. 
% \jun{We don't have participants from other race}

The highest level of education obtained were community college (3.1\%), undergraduate degrees (53.1\%), Masters degrees (25.0\%), medical degrees (15.6\%), and doctorate degrees (3.1\%). The experience levels of participants varied, though a large majority had over 5 years of experience in clinical medicine (12.5\% with 2-5 years, 37.5\% with 5-10 years, 40.6\% with 10-20 years, and 9.4\% with $>$20 years).
Most survey participants are registered nurses or nurse practitioners (78.1\%), with others who identified as doctors/physicians (9.4\%), researchers (6.3\%), or other (6.3\%).

\subsection*{Summary Findings} 
\label{sec:Statistics}
\paragraph{What Clinicians Value in AI} 
Clinicians were most likely to consider \textit{safety} and \textit{performance} important. Clinicians without AI experience selected \textit{privacy} more often, while those with AI experience cared more about \textit{beneficence}. \textit{Accountability}, \textit{human autonomy} and \textit{transparency} were also rated as relatively important among all participants. 
% No participant selected \textit{solidarity}.

\begin{figure}[t!]
    \centering
    \includegraphics[width=0.8\textwidth] {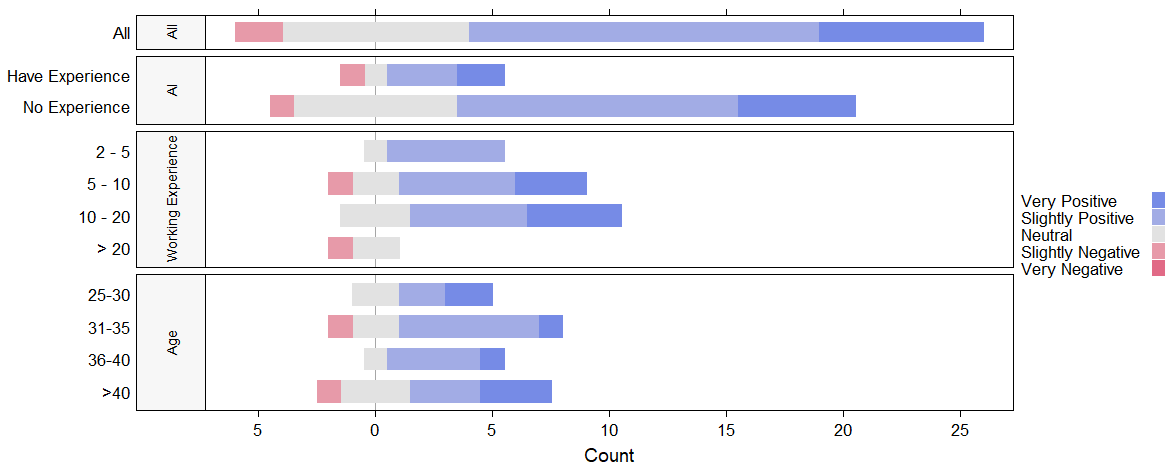}
    \caption{Participant attitudes toward AI by demographic group. }
    \label{fig:Combined_Attitude_Plot} 
\end{figure}

\paragraph{Attitudes Towards AI} 
% We asked participants to report their overall attitude towards AI (answers on a Likert scale with 1 as Most negative and 5 as Most positive). 
Figure~\ref{fig:Combined_Attitude_Plot} reports participant self-reported attitudes towards AI, collectively and split into different demographic groups. Overall distributions in attitudes toward AI are similar across groups. Very experienced clinicians ($>$20 years experience) in our cohort do not exhibit any positive sentiments towards AI; however, we note the small sample size (n=3).

\begin{figure}[t!]
    \centering
    \subfigure[Understandable and reasonable quantified by Likert scale question]{
        \includegraphics[width=0.64\textwidth]{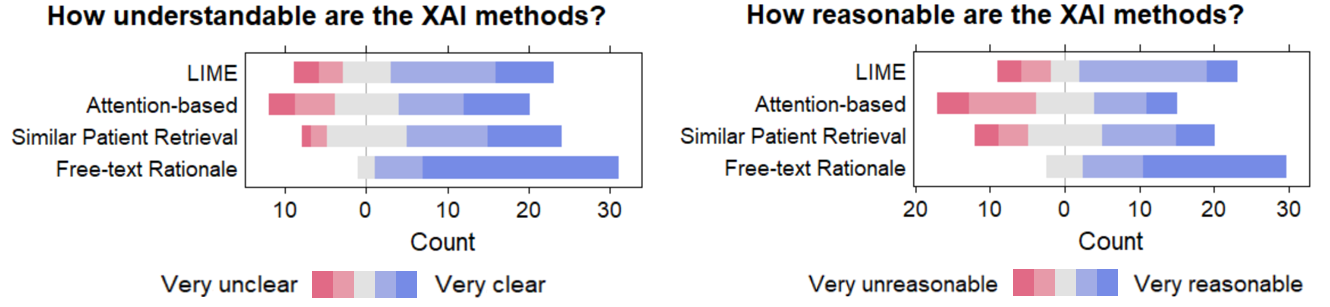}}
        \vspace{2mm}
    \subfigure[Heatmap for Functionality]{
        \includegraphics[width=0.32\textwidth]{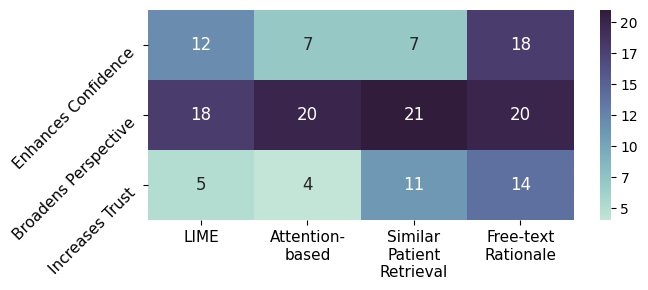}}
    \vspace{-2mm}
    \caption{Overview of practitioners' evaluations on the effectiveness and utility of XAI methods, covering understandability, reasonability, and key functional goals.}
    \label{fig:likert_heatmap}
\end{figure}

\paragraph{Attitudes Toward XAI Methods} 
Regarding how understandable and reasonable each XAI method is from the clinician's perspective (Figure~\ref{fig:likert_heatmap}(a)), we observe similar results for the four XAI methods. Free-text rationales were found to be the most understandable and reasonable, with no negative responses. 
Attention-based explanations were found to be the least reasonable and hardest to understand. LIME and similar patient retrieval received a similar amount of positive and negative feedback along both dimensions, though similar patient retrieval was rated as more understandable. Clinician preferences toward the quantity of information provided, such as the percent of words highlighted or the number of similar patients and rationales shown, generally lean towards preferring more information.
% , but are inconsistent across participants.

\paragraph{Practitioner Preference} 
Overall, free-text rationales were the most preferred method (n=15), with LIME second (n=12). When considering all four methods, no participant ranked Attention-based explanations first.
% were rated as the hardest to understand, with no participant ranking it first. 
The understandability of the similar patient retrieval method showed an almost even distribution across ranks 1-4. However, this method was the least preferred for use in real clinical settings, with an average ranking of 4.

We ask clinicians to assess the time efficiency of their most preferred XAI method on a Likert scale with 1 as least efficient and 5 as most efficient 
%\lucy{instead of most negative and most positive, replace with least efficient and most efficient. otherwise, it's hard to know what's positive and what's negative}. 
For free-text rationales, which is ranked first 15 times, 13\% of participants reported \textit{Minimal time saved}, 20\% \textit{Moderate time saved}, 47\% \textit{Considerable time saved}, and 20\% \textit{Significant time saved}. For LIME, ranked first 12 times, the time savings in the same four categories were assessed as 0\%, 25\%, 67\% and 8\%, respectively. 
In other words, participants were more divided on the time saving ability of free-text rationales.

Figure~\ref{fig:likert_heatmap}(b) visualizes responses towards whether each XAI method succeeds in enhancing confidence, broadening perspective, and increasing trust. The objective of broadening the users' perspective was successfully met by all methods, with more than half of participants agreeing for each method. Enhancing confidence was most effectively addressed by free-text rationales, with 18 participants agreeing, followed by LIME, with 12 participants agreeing. Increasing trust was the most challenging goal, though many more participants indicated similar patient retrieval and free-text rationales as helping to meet this goal. Additional statistics can be found on our Github.

\subsection*{Thematic Analysis} \label{sec:Thematic}
 
\begin{figure}[t!]
    \centering
    \subfigure[Positive sentiment]{%
        \includegraphics[width=0.45\textwidth]{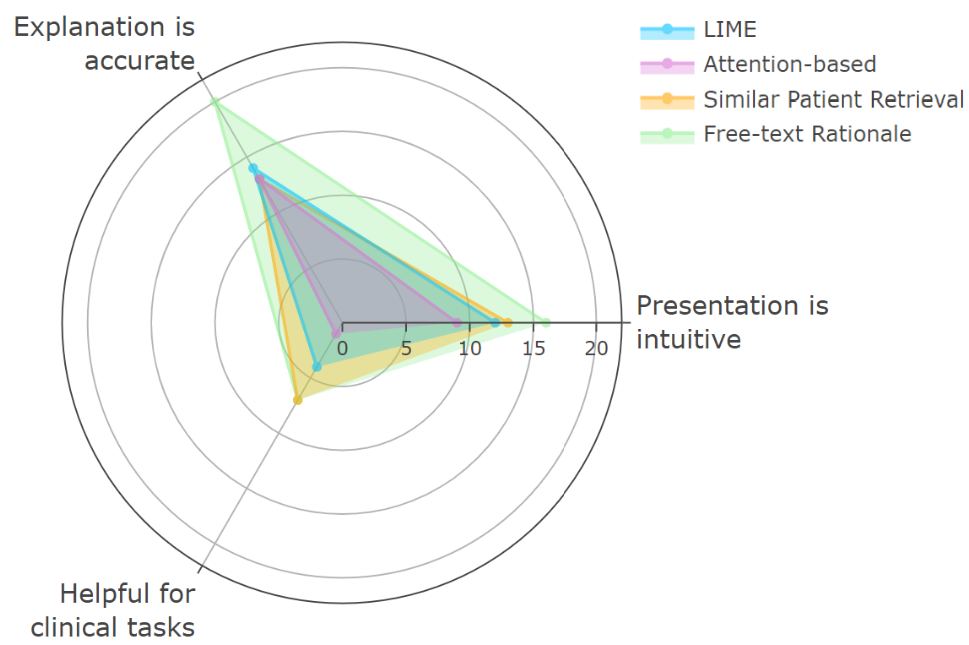}
        \label{fig:thematic_positive}
    }
    \subfigure[Negative sentiment]{%
        \includegraphics[width=0.45\textwidth]{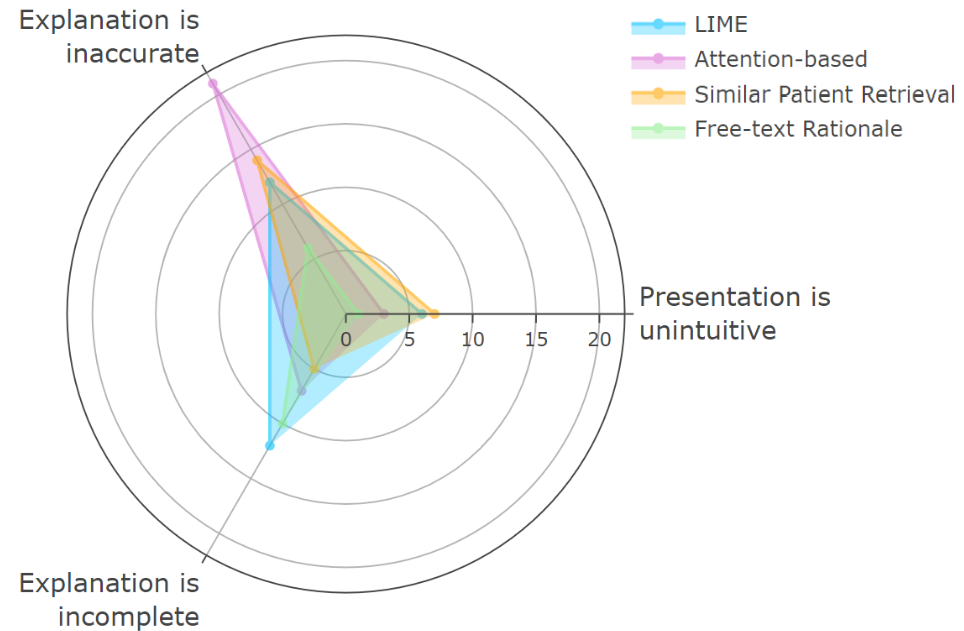}
        \label{fig:thematic_negative}
    }
    \caption{Radar plots of themes mentioned by participants for each XAI method, grouped by sentiment.}
    \label{fig:thematic_radar}
\end{figure}

We conduct qualitative coding to identify shared themes among participant attitudes towards XAI methods.
The first author conduct open coding on survey responses to identify major themes, with feedback and iteration from other authors. This process led to the identification of 6 themes, which we present and describe in Table~\ref{table:sub-themes}, organized by positive and negative sentiment.  Radar plots in Figure~\ref{fig:thematic_radar} compare the number of participants who raised each theme for each XAI method. 
Below, we summarize the qualitative feedback for each XAI method included in our study. Participants are identified by pseudonyms P1-32.

\begin{table}[t!]
\footnotesize
\centering
\begin{tabular}{lL{35mm}L{88mm}}
\toprule
\textbf{Sentiment} & \textbf{Theme} & \textbf{Description} \\
% \midrule
% \multicolumn{2}{l}{\textbf{Positive}} \\
\midrule
Positive & Presentation is intuitive  &   The output of the XAI techniques (visualizations or text) is presented in a way that is intuitive and easy to understand \\
% , enabling the user to quickly understand AI decisions\\
 & Explanation is accurate &  Highlighted keywords and/or free-form explanations align with clinicians' expectations and thought processes \\
 & Helpful for clinical tasks & Explanations are useful in clinical settings, and could assist with tasks like decision-making or patient communication \\
\midrule
% \multicolumn{2}{l}{\textbf{Negative}} \\
% \midrule
Negative & Presentation is unintuitive & The output of the XAI techniques (visualizations or text)
% Visualization techniques 
require further clarification or explanation to be understood by clinicians \\
 & Explanation is inaccurate & Explanations are irrelevant or do not align with clinicians' expectations \\
 & Explanation is incomplete & Additional information is required to fully support decision-making processes \\
\bottomrule
\end{tabular}
\caption{Themes grouped by sentiment with descriptions.
% \lucy{i changed the theme names; please review} \jun{Updated the graph}
}
\label{table:sub-themes}
\end{table}

\paragraph{LIME} 
% As shown in Figure~\ref{fig:thematic_radar}(a), t
The visualization technique of LIME is favored by 12 practitioners, with reasons including \textit{``shade intensity was very helpful in showing how important the phrase was''} (P7) and \textit{``can be helpful in drawing quick conclusions''} (P4). The use of orange and blue for negative and positive features was described as intuitive by 6 practitioners. 
Highlighted words were found to match the clinician's thought process (14 participants); P15 mentioned \textit{``highlighting words such as `intubated' and `unresponsive' is good.''}, and P17 confirmed, \textit{``I found favorable that the system considered more influential (darker shade) the age of the patient.''} P30 offered one potential use of LIME in clinical settings by expressing that \textit{``The red highlighted words did correlate with a mortality risk and were helpful in identifying these risks in the text.''}
Despite generally positive feedback,
some highlighted words were considered not correlated with the outcome (P8) or were located in regions of the admission notes considered irrelevant for prediction (P21 and P28). Furthermore, several participants (P8 and P15) expressed a desire for more unique span highlights rather than duplicates of already highlighted words.
% that contribute to the outcome that are not duplicates of already highlighted words.
% , such as the span `high blood pressure in the 230s.'

\paragraph{Attention-based Highlights}
The attention-based method received the most feedback for the need to improve the quality of highlights, with 21 participants mentioning the low relevancy of highlights to the outcome. Several participants were especially bothered by highlighted words like `of' or `with,' 
described by P12 as ``random'' and P22 as ``distracting from the outcome.''
% commenting that the method \textit{``pick[ed] slightly random words especially in the survived case''} and P22 wanting \textit{``less highlighted words distracting from the outcome.''} 
P2 discussed potential benefits, such as the model \textit{``highlighting important words such as `cardiac' and `received CPR' because my brain thinks those are important too,''} suggesting areas where the model performs well in identifying relevant information.
Regarding visualization, the single-color highlights were favored by 9 participants for their simplicity.
P13, as the only participant commenting on utility for clinical tasks, mentioned that the highlights \textit{``contribute to the respective viewer taking a Quick Look back to make sure no information is missed,''} highlighting the potential of visualization techniques to aid in clinical decision-making.

\paragraph{Similar Patient Retrieval}
Similar patient retrieval was found to be clinically helpful by 7 clinicians:
% as shown in Figure~\ref{fig:thematic_radar}: 
for administrative tasks (P3), planning of treatment (P4, P19, P25), supporting diagnosis (P9, P12, P19), and \textit{``guiding and auditing manifestation and intervention''} (P11). P2 commented that the highlighted NER terms aligned with their thought process, while
% \textit{``as I have seen this in my own experience.''} 
P17 discussed accurate retrieval as a valuable feature of this technique, 
% The accuracy of retrieval, demonstrated by shared valuable features for P17, is notable, 
especially in terms of \textit{``similar age (elderly),... abnormal electrocardiogram signs''} and other shared treatments and symptoms. However, additional clarification in terminology is suggested by P28, who noted \textit{``The use of short hand/abbreviations should be minimized. As this could lead to confusion.''} P27 discusses how this could be used to improve retrieval relevance: \textit{``Consistent use of the same language...
% as opposed to varying 
(eg s/p cardiac arrest vs s/p PEA arrest) 
would help to pull similar patients.''}

Regarding visualizing NER overlap, 9 participants liked this feature, 
% e.g.,
% ability to retrieve similar patients, such as 
% P9 says the highlights 
which allows for \textit{``Quickly referencing similarities in past medical history and treatment''} (P9). 
% Color choices in highlighting were deemed valuable by P1, who 
P1 says \textit{``color related to higher match is helpful.''} P14 suggests the potential for color to convey more detailed categories, e.g., \textit{``Meds: orange, Procedures: yellow, vitals: purple,''} to 
% be more beneficial for 
assist with comparison. However, there is room for improvement in the quality of explanations (14 participants), such as replacing non-contributory medication highlights with more influential words (P16), and wanting higher similarity between queries and retrievals, especially in areas like chief complaints.

\paragraph{Free-text Rationales}
% Figure~\ref{fig:thematic_radar} indicates that the 
Free-text rationales received the most positive feedback,
% were well-received, 
with 16 participants indicating that its outputs were intuitive and 15 indicating that the explanations were accurate. 
% For example, P17 finds \textit{``very valuable the organization in which the explanation was presented.''}
% Reasonable r
Rationales were found to enhance predictability through conciseness of presentation (P16, P17, P19) and accurate reasoning (P23, P24). Participants indicate rationales can aid in clinical tasks such as prognosis (P11, P17, P30), prioritization (P13), decision-making (P15, P19), and treatment planning (P27), offering clarity for understanding (P11, P21) and facilitating communication with non-experts like patients (P12, P30). For improvement, participants suggested considering more contextual details like \textit{``high doses of pressors in the mor[t]ality rate''} (P1) and 
% a more rational ranking, 
the strength of the rationale, e.g., \textit{``the medication allergy reasoning was weak''} (P8). Incorporating quantifiable scores (P7)
% , such as \textit{``quantify risk (i.e. high risk for mortality)''}(P7), 
and \textit{``evidence-based protocols in the rationales, e.g. AHA''} (P11) could provide further 
% refine 
support for outcome predictions, with P21 suggesting that \textit{``Adding confidence levels or percentages would significantly improve how trustworthy this algorithm is.''} 
For presentation enhancements, P31 suggests \textit{``Bold faced and highlighted words for important info''}. On the other hand, use of jargon 
% like `resuscitation' 
and excessive reasoning were found to reduce explanation clarity (P4).

\section*{\textsc{Discussion}} \label{sec:discussion}
\vspace{-1mm}
Our analysis of survey results raises questions about the goals and benefits of XAI methods, and how to increase their relevance and utility to clinical practitioners. We synthesize these findings into additional recommendations below.

\paragraph{The importance of providing evidence}
Similar patient retrieval, while critiqued for its accuracy and effectiveness,
demonstrated a greater ability to build trust than feature-based methods such as LIME and Attention-based highlights. Using historical cases as evidence to support decisions closely mirrors how practitioners rely on past experiences in clinical decision-making. To be successful from the practitioners' perspective, our findings suggest the importance of not only generating accurate and informative rationales but also incorporating evidence-based support (as with the exemplars in similar patient retrieval) in model explanations. While free-text rationales were received positively by participants, the lack of grounded evidence needs to be considered. Combining free-text rationales with retrieved exemplars or externally retrieved evidence (as in Naik et al.\cite{naik-etal-2022-literature}) could help address these issues in future work.
% or tracking features.

\paragraph{Potential of free-text rationales to bridge communication}
Explanations in natural language that reflect the cognitive processes of clinicians can serve as a communication bridge within the healthcare system. 
P27 mentioned that 
in time-sensitive scenarios, generated content can function similarly to a nurse's note for communicating with a doctor. Additionally, generated rationales can be an educational tool, e.g., P27 described \textit{``Especially in the trauma setting, the workflow is very fast, and you got residents attending...even if it's not a teaching hospital, it is extremely helpful.''}
The method also has the potential to bridge the gap between healthcare providers and patients by explaining symptoms and treatments in a non-technical way, as mentioned by P17, \textit{``this model...it would be a good thing to not maybe show the family members, but to explain, okay, we use this model and this is what the outcomes are saying.''}
However, based on participant suggestions to simplify wording in free-text responses, future work should consider integrating plain language summarization\cite{Guo2020AutomatedLL,Guo2023APPLSEE} to enhance the understandability and efficiency of LLM output. 

\paragraph{XAI for both structured and unstructured data}
In the critical care setting, especially in ICUs, treatment plans and bedside monitoring rely heavily on both structured data (such as vital signs and lab results) and unstructured data (such as clinical notes). 
Several participants mentioned the need to incorporate multimodal data into an XAI-enhanced CDSS. While this is beyond the scope of our study, we emphasize that real-world CDSS would likely take advantage of input predictors beyond the clinical note text, and that the importance of these predictors would also require explanation.
This could be an addition to ongoing research that aims to create predictive frameworks combining unstructured textual data with structured data for clinical prediction.\cite{ebrahimi2023lanistr}
% However, there is still room for the evolution of XAI methods capable of handling hybrid data.

\paragraph{XAI tailored to different clinical workflows}
Responses from clinical practitioners revealed varied perceptions and preferences for XAI methods at different stages of patient care. In urgent care settings like the ICU or surgery, clinicians prioritize efficiency and clarity of explanations, favoring XAI methods that present key information quickly for reference. However, in less acute phases like post-surgical care, detailed explanations are in demand for analysis and treatment planning, a role well-served by a method like free-text rationales. 
For example, P17 states, \textit{``Highlight saves time and we need that.~If we had more time in clinical settings, I feel like the free text rationale gives a more in-depth reasoning.''}
To balance details and efficiency,\cite{Jung2023EssentialPA} XAI methods must be tailored to specific clinical workflows.

\paragraph{Limitations} Although we have covered all categories of post-hoc XAI methods from Chaddad et al.,\cite{Chaddad_2023} there are many methods that we could not include in this survey due to time and resource constraints (our survey was already long). Applying XAI methods to text data was not always straightforward, as some methods (like LIME) work better on feature-based models with lower-dimensional input. For similar patient retrieval in particular, we face many challenges in fine-tuning the retrieval model due to the lack of labeled datasets for patient semantic similarity search. 

% We have organized our discussion to maximize the generalizability of our findings to other XAI techniques, though future work could explore and provide  comparative studies of such methods applied to specific clinical tasks.

We use only the free-text admission notes from MIMIC-III as the inputs to our prediction models, whereas prediction models and XAI methods can be applied to other data formats, such as tabular and time series data. Furthermore, the predictive task in this study is limited to in-hospital mortality prediction. Future work should explore multimodal outcome prediction models as well as other clinical predictive tasks.

Although we have made significant efforts to recruit, our cohort is still relatively small, involving 32 clinical practitioners who are predominantly nurses. This may limit the generalizability of our findings. However, we did achieve a fairly diverse cohort in terms of age and experience, and many themes were consistently raised by most participants. In the future, we aim to validate our findings in real-world deployments, which we believe will offer valuable perspectives. We also plan to explore whether clinicians can effectively use XAI to identify hallucinations in LLM-powered decision support and mitigate the risks introduced by such systems.

\subsection*{\textsc{Conclusion}}
\vspace{-1mm}
Our survey results reveal the demands and preferences of healthcare practitioners towards the implementation of XAI in CDSS. By integrating clinicians in the evaluation process, we observed a strong preference for XAI techniques that replicate clinical reasoning, such as exemplar-based patient retrieval and free-text rationales. These methods enhance the interpretability and trustworthiness of AI-supported decision making, which can further help in realize the full potential of AI in clinical decision making, ensuring that CDSS are not only effective but also align with healthcare providers' needs. Moving forward, we aim to refine these methods by incorporating structured and unstructured data, tailoring XAI approaches to specific clinical workflows. This will improve the utility and efficacy of CDSS across diverse clinical settings, further supporting healthcare professionals in their decision-making processes.

\section*{\textsc{Acknowledgements}} 
We acknowledge support from the University of Washington Institute for Medical Data Science and the eScience Institute's Azure Cloud Credits for Research and Teaching.

% % References as numbers
% \makeatletter
% \renewcommand{\@biblabel}[1]{\hfill #1.}
% \makeatother

% unstr is used to keep citation order
%\todoit{a lot of these references are to arxiv papers when there is a better version of record from ACL or ACM. i already replaced a few of them. could you do the rest?}
\bibliographystyle{unsrt}
\bibliography{sample}  

\begin{thebibliography}{10}

\bibitem{danilevsky2020survey}
Marina Danilevsky, Kun Qian, Ranit Aharonov, Yannis Katsis, Ban Kawas, and Prithviraj Sen.
\newblock A survey of the state of explainable {AI} for natural language processing.
\newblock In {\em Proceedings of the 1st Conference of the Asia-Pacific Chapter of the Association for Computational Linguistics}, pages 447--459, Suzhou, China, December 2020.

\bibitem{feng2020explainable}
Jinyue Feng, Chantal Shaib, and Frank Rudzicz.
\newblock Explainable clinical decision support from text.
\newblock In {\em Proceedings of the 2020 conference on empirical methods in natural language processing (EMNLP)}, pages 1478--1489, 2020.

\bibitem{Sheikhalishahi_2019}
Seyedmostafa Sheikhalishahi, Riccardo Miotto, Joel~T Dudley, Alberto Lavelli, Fabio Rinaldi, and Venet Osmani.
\newblock Natural language processing of clinical notes on chronic diseases: Systematic review.
\newblock In {\em JMIR Medical Informatics}, page e12239. JMIR Publications Inc., 2019.

\bibitem{Chaddad_2023}
A.~Chaddad, J.~Peng, J.~Xu, and A.~Bouridane.
\newblock Survey of explainable {AI} techniques in healthcare.
\newblock In {\em Sensors (Basel, Switzerland)}, page 634. MDPI, 2023.

\bibitem{di2023explainable}
Flavio Di~Martino and Franca Delmastro.
\newblock Explainable ai for clinical and remote health applications: a survey on tabular and time series data.
\newblock {\em Artificial Intelligence Review}, 56(6):5261--5315, 2023.

\bibitem{Ribeiro2016WhySI}
Marco~Tulio Ribeiro, Sameer Singh, and Carlos Guestrin.
\newblock “{Why Should I Trust You?}”: Explaining the predictions of any classifier.
\newblock {\em Proceedings of the 22nd ACM SIGKDD International Conference on Knowledge Discovery and Data Mining}, 2016.

\bibitem{Vaswani2017AttentionIA}
Ashish Vaswani, Noam~M. Shazeer, Niki Parmar, Jakob Uszkoreit, Llion Jones, Aidan~N. Gomez, Lukasz Kaiser, and Illia Polosukhin.
\newblock Attention is all you need.
\newblock In {\em Neural Information Processing Systems}, 2017.

\bibitem{van-aken-etal-2021-clinical}
Betty van Aken, Jens-Michalis Papaioannou, Manuel Mayrdorfer, Klemens Budde, Felix Gers, and Alexander Loeser.
\newblock Clinical outcome prediction from admission notes using self-supervised knowledge integration.
\newblock In {\em Proceedings of the 16th Conference of the European Chapter of the Association for Computational Linguistics: Main Volume}, pages 881--893, April 2021.

\bibitem{jin2018improving}
Mengqi Jin, Mohammad Bahadori, Aaron Colak, Parminder Bhatia, Busra Celikkaya, Ram Bhakta, Selvan Senthivel, Mohammed Khalilia, Daniel Navarro, Borui Zhang, Tiberiu Doman, Arun Ravi, Matthieu Liger, and Taha Kass-hout.
\newblock Improving hospital mortality prediction with medical named entities and multimodal learning, 2019.

\bibitem{lundberg2017unified}
Scott~M Lundberg and Su-In Lee.
\newblock A unified approach to interpreting model predictions.
\newblock {\em Advances in neural information processing systems}, 30, 2017.

\bibitem{garg2023annotated}
Muskan Garg, Amirmohammad Shahbandegan, Amrit Chadha, and Vijay Mago.
\newblock An annotated dataset for explainable interpersonal risk factors of mental disturbance in social media posts.
\newblock In {\em Findings of the Association for Computational Linguistics: ACL 2023}, pages 11960--11969, Toronto, Canada, July 2023.

\bibitem{Saxena2022ExplainableCA}
Chandni Saxena, Muskan Garg, and Gunjan Saxena.
\newblock Explainable causal analysis of mental health on social media data.
\newblock In {\em Neural Information Processing}, pages 172--183, Cham, 2023. Springer International Publishing.

\bibitem{Agerri2023HiTZAntidoteAE}
Rodrigo Agerri, I{\~n}igo Alonso, Aitziber Atutxa, Ander Berrondo, Ainara Estarrona, Iker Garc{\'i}a-Ferrero, Iakes Goenaga, Koldo Gojenola, Maite Oronoz, Igor Perez-Tejedor, German Rigau, and Anar Yeginbergenova.
\newblock Hitz@antidote: Argumentation-driven explainable artificial intelligence for digital medicine.
\newblock In {\em Annual Conference of the Spanish Society for Natural Language Processing}, 2023.

\bibitem{Wang2023CanLL}
Zhuo Wang, Rong~Hua Li, Bowen Dong, Jie Wang, Xiuxing Li, Ning Liu, Chenhui Mao, Wei Zhang, Liling Dong, Jing Gao, and Jianyong Wang.
\newblock Can {LLMs} like {GPT-4} outperform traditional ai tools in dementia diagnosis? maybe, but not today.
\newblock {\em ArXiv}, abs/2306.01499, 2023.

\bibitem{shen2018constructing}
Ying Shen, Jo{\"e}l Colloc, Armelle Jacquet-Andrieu, Ziyi Guo, and Yong Liu.
\newblock Constructing ontology-based cancer treatment decision support system with case-based reasoning.
\newblock In {\em Smart Computing and Communication: Second International Conference, SmartCom 2017, Shenzhen, China, December 10-12, 2017, Proceedings 2}, pages 278--288. Springer, 2018.

\bibitem{Markus_2021}
Aniek~F. Markus, Jan~A. Kors, and Peter~R. Rijnbeek.
\newblock The role of explainability in creating trustworthy artificial intelligence for health care: A comprehensive survey of the terminology, design choices, and evaluation strategies.
\newblock In {\em Journal of Biomedical Informatics}, page 103655. Elsevier BV, 2021.

\bibitem{Marafino_2018}
B.~J. Marafino, M.~Park, J.~M. Davies, R.~Thombley, H.~S. Luft, D.~C. Sing, D.~S. Kazi, C.~DeJong, W.~J. Boscardin, M.~L. Dean, and R.~A. Dudley.
\newblock Validation of prediction models for critical care outcomes using natural language processing of electronic health record data.
\newblock In {\em JAMA Network Open}, page e185097, 2018.

\bibitem{naik-etal-2022-literature}
Aakanksha Naik, Sravanthi Parasa, Sergey Feldman, Lucy~Lu Wang, and Tom Hope.
\newblock Literature-augmented clinical outcome prediction.
\newblock In {\em Findings of the Association for Computational Linguistics: NAACL 2022}, pages 438--453, Seattle, United States, July 2022. Association for Computational Linguistics.

\bibitem{antoniadi2021current}
Anna~Markella Antoniadi, Yuhan Du, Yasmine Guendouz, Lan Wei, Claudia Mazo, Brett~A Becker, and Catherine Mooney.
\newblock Current challenges and future opportunities for xai in machine learning-based clinical decision support systems: a systematic review.
\newblock {\em Applied Sciences}, 11(11):5088, 2021.

\bibitem{Jung2023EssentialPA}
Jinsun Jung, Hyung-Mok Lee, Hyunggu Jung, and Hyeoneui Kim.
\newblock Essential properties and explanation effectiveness of explainable artificial intelligence in healthcare: A systematic review.
\newblock {\em Heliyon}, 9, 2023.

\bibitem{Johnson_2016}
A.~Johnson, T.~Pollard, and L.~et~al. Shen.
\newblock Mimic-iii, a freely accessible critical care database.
\newblock In {\em Scientific Data}, page 160035, 2016.

\bibitem{michalopoulos-etal-2021-umlsbert}
George Michalopoulos, Yuanxin Wang, Hussam Kaka, Helen Chen, and Alexander Wong.
\newblock {U}mls{BERT}: Clinical domain knowledge augmentation of contextual embeddings using the {U}nified {M}edical {L}anguage {S}ystem {M}etathesaurus.
\newblock In {\em Proceedings of the 2021 Conference of the North American Chapter of the Association for Computational Linguistics: Human Language Technologies}, pages 1744--1753, Online, June 2021.

\bibitem{falaki2023attention}
Ala~Alam Falaki and Robin Gras.
\newblock Attention visualizer package: Revealing word importance for deeper insight into encoder-only transformer models, 2023.

\bibitem{Reimers2019SentenceBERTSE}
Nils Reimers and Iryna Gurevych.
\newblock Sentence-bert: Sentence embeddings using siamese bert-networks.
\newblock In {\em Conference on Empirical Methods in Natural Language Processing}, 2019.

\bibitem{neumann-etal-2019-scispacy}
Mark Neumann, Daniel King, Iz~Beltagy, and Waleed Ammar.
\newblock {S}cispa{C}y: Fast and robust models for biomedical natural language processing.
\newblock In {\em Proceedings of the 18th BioNLP Workshop and Shared Task}, pages 319--327, Florence, Italy, August 2019. Association for Computational Linguistics.

\bibitem{openai2024gpt4}
OpenAI.
\newblock {GPT-4} technical report, 2024.

\bibitem{liu-etal-2022-makes}
Jiachang Liu, Dinghan Shen, Yizhe Zhang, Bill Dolan, Lawrence Carin, and Weizhu Chen.
\newblock What makes good in-context examples for {GPT}-3?
\newblock In {\em Proceedings of Deep Learning Inside Out (DeeLIO 2022)}, pages 100--114, Dublin, Ireland and Online, May 2022. Association for Computational Linguistics.

\bibitem{brauner2023does}
Philipp Brauner, Alexander Hick, Ralf Philipsen, and Martina Ziefle.
\newblock What does the public think about artificial intelligence?—a criticality map to understand bias in the public perception of {AI}.
\newblock {\em Frontiers in Computer Science}, 5:1113903, 2023.

\bibitem{kaya2024roles}
Feridun Kaya, Fatih Aydin, Astrid Schepman, Paul Rodway, Okan Yeti{\c{s}}ensoy, and Meva Demir~Kaya.
\newblock The roles of personality traits, {AI} anxiety, and demographic factors in attitudes toward artificial intelligence.
\newblock {\em International Journal of Human--Computer Interaction}, 40(2):497--514, 2024.

\bibitem{jakesch2022different}
Maurice Jakesch, Zana Bu{\c{c}}inca, Saleema Amershi, and Alexandra Olteanu.
\newblock How different groups prioritize ethical values for responsible {AI}.
\newblock In {\em Proceedings of the 2022 ACM Conference on Fairness, Accountability, and Transparency}, pages 310--323, 2022.

\bibitem{saeed2023explainable}
Waddah Saeed and Christian Omlin.
\newblock Explainable {AI (XAI)}: A systematic meta-survey of current challenges and future opportunities.
\newblock {\em Knowledge-Based Systems}, 263:110273, 2023.

\bibitem{Guo2020AutomatedLL}
Yue Guo, Weijian Qiu, Yizhong Wang, and Trevor~A. Cohen.
\newblock Automated lay language summarization of biomedical scientific reviews.
\newblock {\em AAAI}, 2021.

\bibitem{Guo2023APPLSEE}
Yue Guo, Tal August, Gondy Leroy, Trevor~A. Cohen, and Lucy~Lu Wang.
\newblock {APPLS}: Evaluating evaluation metrics for plain language summarization.
\newblock {\em arXiv}, 2023.

\bibitem{ebrahimi2023lanistr}
Sayna Ebrahimi, Sercan~O. Arik, Yihe Dong, and Tomas Pfister.
\newblock {LANISTR}: Multimodal learning from structured and unstructured data, 2023.

\end{thebibliography}

% \newpage
% \appendix

% \input{appendix}

\end{document}